\documentclass[10pt,twocolumn,letterpaper]{article}

\usepackage{cvpr}              

\usepackage{graphicx}
\usepackage{amsmath}
\usepackage{amssymb}
\usepackage{booktabs}
\usepackage{multirow}
\usepackage{pifont}
\usepackage{xcolor}
\usepackage{stfloats}
\usepackage[pagebackref,breaklinks,colorlinks]{hyperref}

\usepackage[capitalize]{cleveref}
\crefname{section}{Sec.}{Secs.}
\Crefname{section}{Section}{Sections}
\Crefname{table}{Table}{Tables}
\crefname{table}{Tab.}{Tabs.}

\def\cvprPaperID{} 
\def\confName{CVPR}
\def\confYear{2023}

\begin{document}

\title{Side Adapter Network for Open-Vocabulary Semantic Segmentation}

\author{Mengde Xu$^{1,2}\thanks{Equal Contribution}$, Zheng Zhang$^{1,2}$\footnotemark[1], Fangyun Wei$^{2}$, Han Hu$^{2}$, Xiang Bai$^{1}$ \\
\\
{$^1$Huazhong University of Science and Technology} \quad
{$^2$Microsoft Research Asia} \\ \\
\href{https://mendelxu.github.io/SAN}{https://mendelxu.github.io/SAN}
}
\twocolumn[{%
\renewcommand\twocolumn[1][]{#1}%
\maketitle
\begin{center}
    \includegraphics[width=0.95\linewidth]{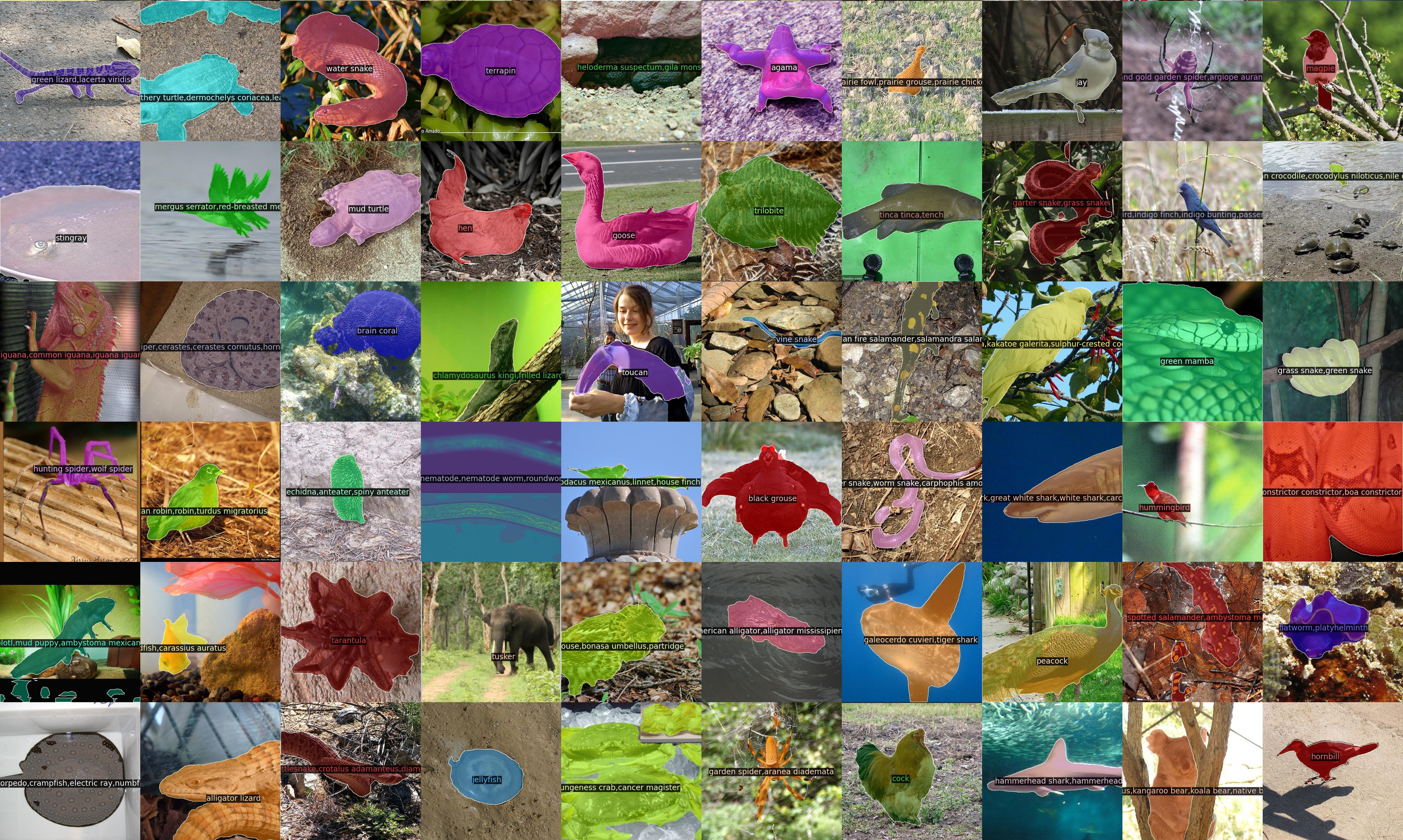}
    \captionof{figure}{Segmentation results on ImageNet. For each image, we combine its category with the coco categories as the vocabulary during inference and only visualize mask of the annotated category.}
    \label{fig:vis_img}
    \vspace{0.5em}
\end{center}%
}]

\maketitle


\begin{abstract}
This paper presents a new framework for open-vocabulary semantic segmentation with the pre-trained vision-language model, named Side Adapter Network (SAN). Our approach models the semantic segmentation task as a region recognition problem. A side network is attached to a frozen CLIP model with two branches: one for predicting mask proposals, and the other for predicting attention bias which is applied in the CLIP model to recognize the class of masks. This decoupled design has the benefit CLIP in recognizing the class of mask proposals. Since the attached side network can reuse CLIP features, it can be very light. In addition, the entire network can be trained end-to-end, allowing the side network to be adapted to the frozen CLIP model, which makes the predicted mask proposals CLIP-aware.
Our approach is fast, accurate, and only adds a few additional trainable parameters. We evaluate our approach on multiple semantic segmentation benchmarks. Our method significantly outperforms other counterparts, with up to 18 times fewer trainable parameters and 19 times faster inference speed.~\cref{fig:vis_img} shows some visualization results on ImageNet. We hope our approach will serve as a solid baseline and help ease future research in open-vocabulary semantic segmentation.
\end{abstract}

\section{Introduction}
\label{sec:intro}

Recognizing and segmenting the visual elements of any category is the pursuit of semantic segmentation. Modern semantic segmentation methods~\cite{long2015fully,chen2017deeplab,cheng2021mask2former} rely on large amounts of labeled data, but typically datasets often only consist of tens to hundreds of categories, and expensive data collection and annotation limit our possibilities to further expand the categories. Recently, large-scale vision-language models~\cite{radford2021learning,jia2021scaling,florence2021,coca2022}, represented by CLIP~\cite{radford2021learning}, have enabled arbitrary category recognition at the image level, \emph {i.e., open-vocabulary image classification}, and this great success encourages us to explore its adaptation in semantic segmentation.

Applying the CLIP model in open-vocabulary semantic segmentation is challenging because the CLIP model is trained by image-level contrastive learning. Its learned representation lacks the pixel-level recognition capability that is required for semantic segmentation. One solution~\cite{ghiasi2021oseg,li2022language} to remedy the granularity gap of representation is fine-tuning the model on the segmentation dataset. However, the data sizes of segmentation datasets are much less than the vision-language pre-training dataset, so the capability of fine-tuned models on open-vocabulary recognition is often compromised.

Modeling semantic segmentation as a region recognition problem bypasses the above difficulties. Early attempts~\cite{xu2022simple,ding2022decoupling} adopt a two-stage training framework. In the first stage, a stand-alone model is trained to generate a set of masked image crops as mask proposals. In the second stage, the vision-language pre-training model (\eg CLIP) is used to recognize the class of masked image crops. However, since the mask prediction model is completely independent of the vision-language pre-training model, it misses the opportunity to leverage the strong features of the vision-language pre-training model and the predicted masked image crops may be unsuitable for recognition, which leads to a heavy, slow, and low-performing model.

\begin{figure}
    \centering
    \includegraphics[width=0.47\textwidth]{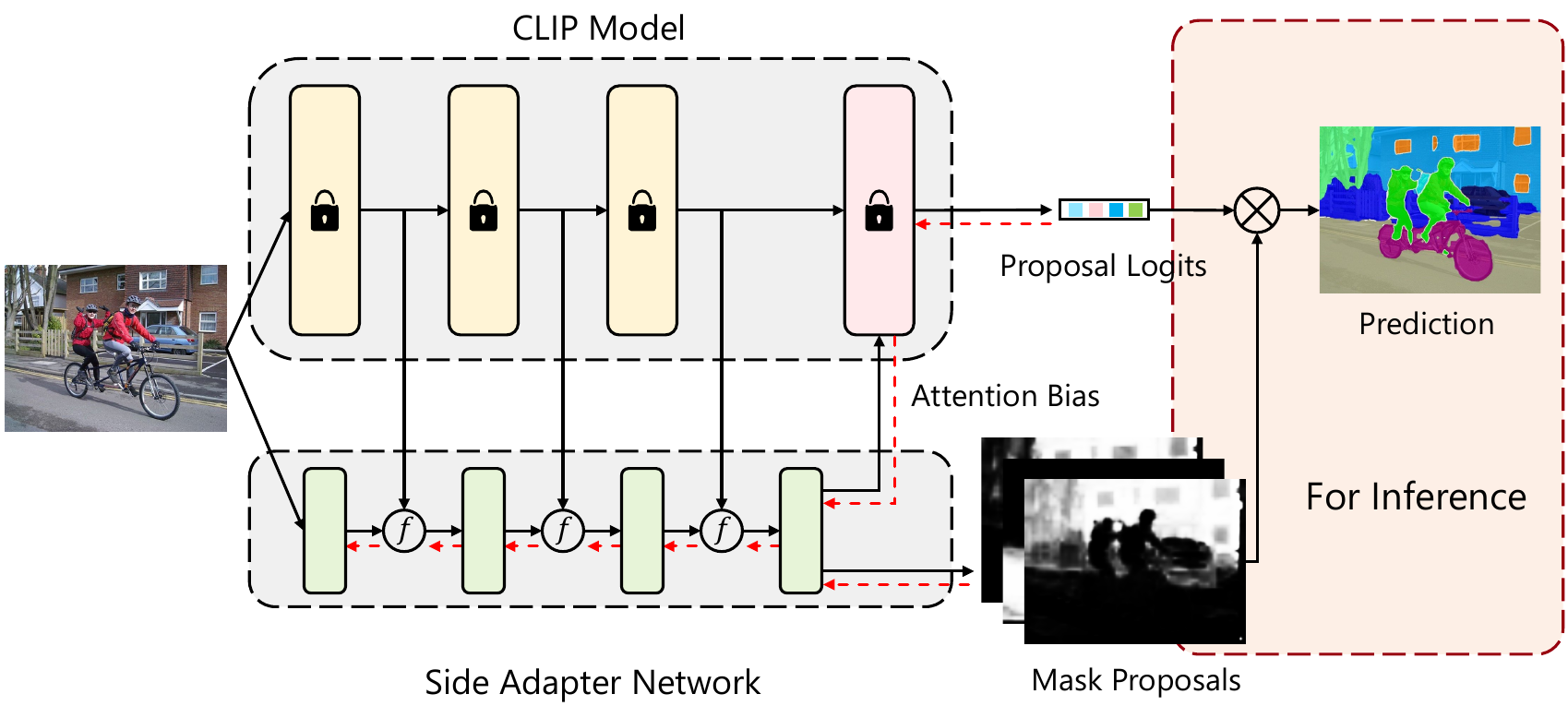}
    \caption{Overview of our \emph{SAN}. The red dotted lines indicate the gradient flow during training. In our framework, the frozen CLIP model still serves as a classifier, and the side adapter network generates mask proposals and attention bias to guide the deeper layers of the CLIP model to predict proposal-wise classification logits. During inference, the mask proposals and the proposal logits are combined to get final predictions through \emph{Matmul}.}
    \label{fig:arch}
\end{figure}

This work seeks to fully unleash the capabilities of the vision-language pre-training model in open vocabulary semantic segmentation. To reach this goal, we present a new framework (\cref{fig:arch}), called \emph{side adapter network} (SAN). Its mask prediction and recognition are \emph{CLIP-aware} because of end-to-end training, and it can be lightweight due to leveraging the features of CLIP.

The side adapter network has two branches: one predicting mask proposals, and one predicting attention biases that are applied to the self-attention blocks of CLIP for mask class recognition. We show this \emph{decoupled} design improves the segmentation performance because the region used for CLIP to recognize the mask may be different from the mask region itself. To minimize the cost of CLIP, we further present a \textit{single-forward} design: the features of shallow CLIP blocks are fused to SAN, and other deeper blocks are combined with attention biases for mask recognition. Since the training is end-to-end, the side adapter network can be maximally adapted to the frozen CLIP model. 

With the aim of fairness and reproducibility, our study is based on officially released CLIP models. We focus on the released ViT CLIP models because the vision transformer has \emph{de facto} substituted ConvNet as the dominant backbone in the computer vision community, and for conceptual consistency and simplicity, the side adapter network is also implemented by the vision transformer.

Accurate semantic segmentation needs high-resolution images, but the released ViT CLIP models are designed for low-resolution images (\eg$224\times224$) and directly apply to high-resolution images giving a poor performance. To alleviate the conflicts in input resolutions, we use low-resolution images in the CLIP model and high-resolution images in the side adapter network. We show this \emph{asymmetric input resolution} is very effective. In addition, we also explore only fine-tuning the positional embedding of the ViT model and note improvements.

We evaluate our method on various benchmarks. Following the setting of previous works~\cite{xu2022simple,liang2022open}, the COCO Stuff~\cite{caesar2018coco} dataset is used for training, and Pascal VOC~\cite{everingham2011pascal}, Pascal Context-59~\cite{mottaghi2014role}, Pascal Context-459~\cite{mottaghi2014role}, ADE20K-150~\cite{zhou2017scene}, and ADE20K-847~\cite{zhou2017scene} are used for testing. Without bells and whistles, we report state-of-the-art performance on all benchmarks: with the CLIP ViT-L/14 model, our method achieves 12.4 mIoU on ADE-847, 15.7 mIoU on PC-459, 32.1 mIoU on ADE-150, 57.7 mIoU on PC-59, and 94.6 mIoU on VOC. Compared to the previous best method, our method has an average of +1.8 mIoU improvements on 5 datasets for ViT-B/16, and +2.3 mIoU improvements for ViT-L/14, respectively. By further applying ensemble trick, the average performance gap increases to +2.9 mIoU and +3.7 mIoU for ViT-B/16 and ViT-L/14.

Along with the excellent performance, our approach requires only 8.4M trainable parameters with 64.3 GFLOPs, which is only $13\%$ and $20\%$ of ~\cite{ding2022open}, $6\%$ and less than $1\%$ of ~\cite{liang2022open}, respectively. 

\begin{figure*}
    \centering
    \includegraphics[width=\textwidth]{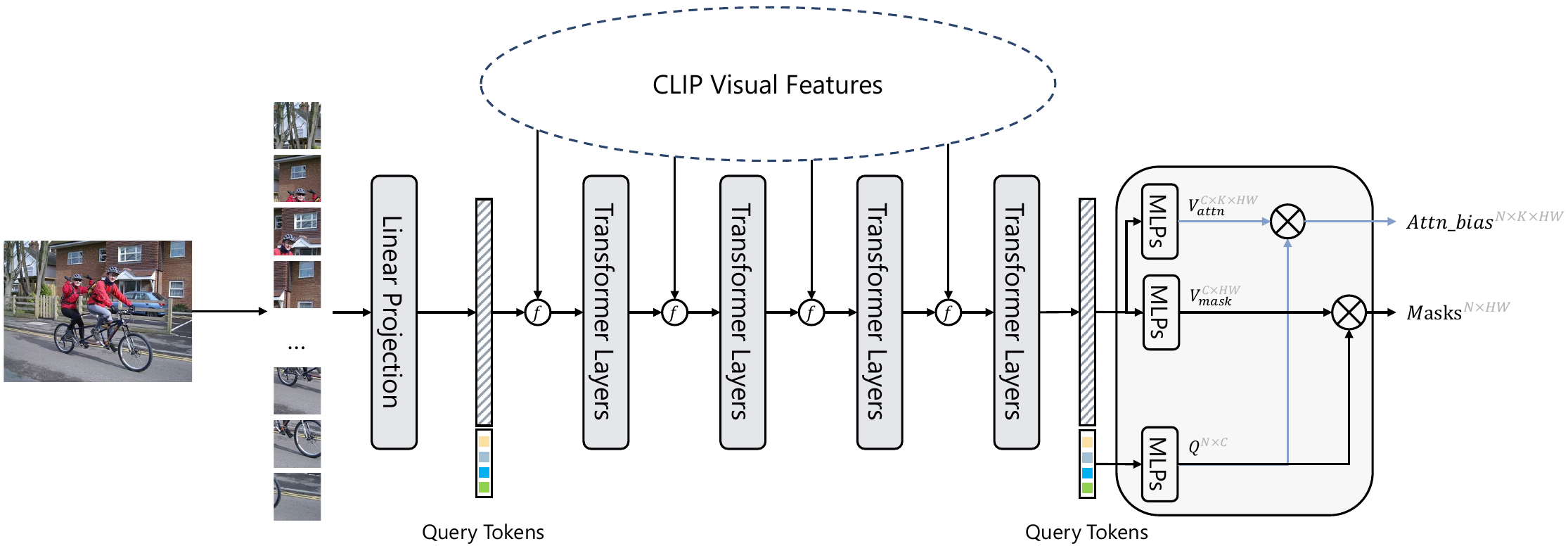}
    \caption{The architecture of the side adapter network. The side adapter network projects the input image to visual tokens and appends query tokens to them at the beginning. Further, it fuses the immediate features of the CLIP model in the middle of transformer layers. The query and visual features are encoded with MLP layers to generate the attention biases and the mask proposals.}
    \label{fig:san}
\end{figure*}
\section{Related Works}
\paragraph{Large-scale vision-language pre-training model}
The goal of visual-language pre-training is to learn generic representations of vision and language. Early works~\cite{su2019vl,lu2019vilbert,chen2019uniter,li2020oscar} in this area mainly followed the paradigm of first pre-training models on visual and language data of moderate size, and then fine-tuning them on downstream visual-language tasks, such as VQA~\cite{antol2015vqa} and image captioning, to validate the benefits of pre-training. Recently, however, CLIP~\cite{radford2021learning} and ALIGN~\cite{jia2021scaling} demonstrates that visual-language models pre-trained on large-scale noisy text-image pairs also have the capabilities on open-vocabulary recognition, in addition to serving as a good starting point for downstream tasks. Many recent works have also confirmed the observation and achieved impressive performance on open-vocabulary image recognition~\cite{florence2021,coca2022,flamingo} and other downstream tasks~\cite{gu2021open, wang2022cris, hessel2021clipscore, patashnik2021styleclip}. 

Our work further explores leveraging the vision-language pre-training models' ability to open-vocabulary recognition on semantic segmentation, which is more challenging with the misalignment between the pre-training and the pixel-level recognition. Specially, we focus on the CLIP model and extend its power in open-vocabulary semantic segmentation.

\paragraph{Models Tuning of downstream tasks.}
Fine-tuning all model parameters is the most common approach to leverage the pre-training models on downstream tasks. However, as pre-training models become larger and stronger, fine-tuning gradually become an inefficient approach and can compromise the model capability learned in the pre-training stage. Therefore, the new approaches to model tuning are starting to attract attention. Earlier explorations~\cite{houlsby2019parameter,lester2021power,li2021prefix} appeared first in NLP community. Recently, with the emergence of large-scale vision models, the exploration in computer vision has also become intensive. CoOp~\cite{zhou2021learning} fine-tunes the CLIP model for image classification tasks by training only the input prompt of the CLIP's text encoder. Tip-Adapter~\cite{zhang2021tip} and VL-Adapter~\cite{sung2022vladapter} insert trainable adapter modules into a fixed CLIP model and finetune only the adapters with few-shot supervision. These methods mainly focus on image-level recognition tasks or vision-language tasks. 

The most related works to us are Side-Tuning~\cite{zhang2020side} and its variants~\cite{sung2022lst}, a side network is attached to the pre-training model and the final representation is a combination of the side network and the pre-training model. However, these efforts are mostly conceptual works and cannot be directly used for open-vocabulary semantic segmentation.

\paragraph{Open-vocabulary Semantic Segmentation}
Earlier work~\cite{zhao2017open,xian2019semantic,bucher2019zero} for open-vocabulary semantic segmentation focus on learning a joint embedding space between image pixels and class name/description. Most recently, driven by the effectiveness of large-scale vision-language pre-training models for open-vocabulary recognition, many approaches explore their application on open-vocabulary semantic segmentation. Some of them~\cite{li2022language,zhou2021denseclip,ghiasi2021oseg,liang2022open} fine-tune the vision-language pre-training models, which requires a large amount of additional data or compromises the open-vocabulary capability of the vision-language pre-training model. 

SimSeg~\cite{xu2022simple} presents a two-stage framework: first generating masked image crops and then recognizing the crops by a frozen CLIP. However, it requires a heavy mask generator, and CLIP must be forwarded multiple times, making it inefficient in terms of both model size and inference speed. Besides, the mask generator is \emph{CLIP-unaware}, further limiting its performance.
MaskCLIP~\cite{ding2022open} improves the two-stage framework by progressively refining the predicted masks by the CLIP encoder, and applying masks in attention layers to avoid forwarding multiple times, which was first introduced by\cite{cheng2021mask2former}. However, MaskCLIP still needs a heavy mask generator, the initial mask prediction is also \emph{CLIP-unaware}, and the mask prediction and recognition are \emph{coupled}. 

Our approach is an end-to-end framework, the mask prediction is lightweight and \emph{CLIP-aware}, and the mask recognition is \emph{decoupled} from mask prediction. These differences allow our approach can better leverage the capability of CLIP than two-stage approaches~\cite{xu2022simple,ding2022open}.

\section{Side Adapter Network}

To fully unleash the capability of CLIP in open vocabulary semantic segmentation, we present \emph{Side Adapter Network} (SAN), which is an end-to-end framework where mask prediction and recognition are intertwined with the CLIP model. The SAN is implemented by a lightweight vision transformer that can leverage the feature of CLIP, and it has two types of outputs: mask proposals and attention biases. The attention biases are applied to the self-attention of CLIP for recognizing the class of mask proposals. In practice, we fuse the feature of shallow CLIP layers into SAN, and apply the attention biases to rest deeper CLIP layers for recognition. With this \textit{single-forward} design, the cost of CLIP model can be minimized. 

The detailed architecture of SAN is shown in~\cref{fig:san}. The input image is split into $16\times16$ patches. A linear embedding layer is applied to project patches as visual tokens. These visual tokens are then concatenated with $N$ learnable query, and fed into subsequent transformer layers. Following the common practices~\cite{cheng2021per,cheng2021mask2former}, we add the absolute position embedding in each transformer block for both visual tokens and query tokens. The position embedding is shared across layers. 

There are two outputs of SAN: the mask proposals and the corresponding attention biases used for mask recognition. In mask prediction, the query tokens and visual tokens are first projected as 256-dimension by two individual 3-layer MLPs, we denoted the projected query tokens as $\mathbf{Q}_{\texttt{mask}} \in \mathbb{R}^{N\times256}$, where $N$\footnote{By default, $N$=100.} is the number of query tokens, and the projected visual tokens as $\mathbf{V}_{\texttt{mask}} \in \mathbb{R}^{\frac{H}{16} \times \frac{W}{16} \times 256}$, where $H$ and $W$ are the height and width of the input image. Then, the masks are generated by the inner product of $\mathbf{Q}_{\texttt{mask}}$ and $\mathbf{V}_{\texttt{mask}}$:
\begin{align}
    \mathbf{M} = \mathbf{V}_{\texttt{mask}} \mathbf{Q}_{\texttt{mask}}^{\mathrm{T}}
\end{align}
, where $\mathbf{M} \in \mathbb{R}^{\frac{H}{16} \times \frac{W}{16} \times N}$.
Generating attention bias is similar to mask prediction. The query tokens and visual tokens are also projected by a 3-layer MLPs, denoted as $\mathbf{Q}_{\texttt{attn}} \in \mathbb{R}^{N\times256}$ and $\mathbf{V}_{\texttt{attn}} \in \mathbb{R}^{\frac{H}{16} \times \frac{W}{16} \times K \times 256}$, where $K$ is the attention head number of ViT CLIP model. By inner producing $\mathbf{Q}_{\texttt{attn}}$ and $\mathbf{V}_{\texttt{attn}}$, we have the the attention biases:
\begin{align}
    \mathbf{B} = \mathbf{V}_{\texttt{attn}} \mathbf{Q}_{\texttt{attn}}^{\mathrm{T}}
\end{align}
,where $\mathbf{B} \in \mathbb{R}^{\frac{H}{16} \times \frac{W}{16} \times K \times N}$. In addition, if needed, the attention biases will be further resized to $\mathbf{B} \in \mathbb{R}^{h \times w \times K \times N}$, where $h$ and $w$ is the height and width of the attention map in CLIP.
In practice, the $\mathbf{Q}_{\texttt{mask}}$ and $\mathbf{Q}_{\texttt{attn}}$ can be shared, and the attention biases will be applied in several self-attention layers of CLIP, \ie the biases are used in different self-attention layers. 

The motivation behind the \emph{decoupled} design of mask prediction and recognition is intuitive: the region of interest used to recognize the mask in CLIP may differ from the mask region itself. We show the effectiveness of this design in~\cref{tab:aba_decouple}.

\paragraph{Feature fusion on visual tokens} The ViT model consists of visual tokens and a \texttt{[CLS]} token, but we only fuse the visual tokens to the SAN.  Since the number and feature dimension of the visual tokens may be different between CLIP and SAN, we first re-arrange visual tokens to feature maps that undergo a $1 \times 1$ convolution and the resize operation to adjust channel dimension and feature map size, and then merged them with the corresponding feature map of SAN by element-wise addition. The feature fusion will be performed several times, taking the 12-layer ViT-B/16 CLIP model and an 8-layers SAN model as an example. We fuse the feature of \{\texttt{stem}, 3, 6, 9\} layer of CLIP with the feature of \{\texttt{stem}, 1, 2, 3\} layer of SAN. 

Our feature fusion has an intuitive design and a more sophisticated structure may improve the performance, but it is not the focus of this work.

\begin{figure}
    \centering
    \includegraphics[width=0.47\textwidth]{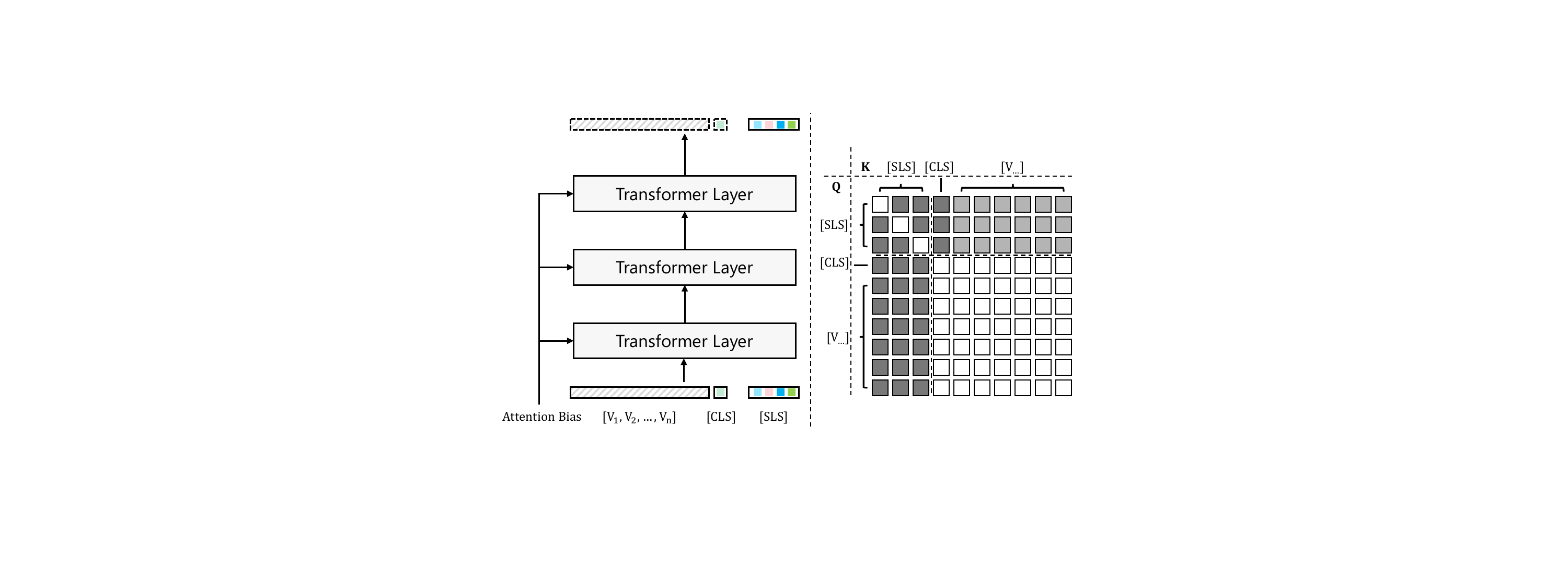}
    \caption{Illustration of using attention bias in CLIP to predict masks. \textbf{(Left)} A set of \texttt{[SLS]} tokens (\ie The \emph{shadow} \texttt{[CLS]} token copies) are created and applied to CLIP. These \texttt{[SLS]} tokens are updated under the effect of attention bias. \textbf{(Right)} The diagram shows how different types of tokens interact with each other. The color of the squares indicates the relationship between query token and key token: black means query is not updated by key, white means query can normally be updated by key, and gray means the the query can be updated by key under the effect of attention bias.}
    \label{fig:head}
\end{figure}

\paragraph{Mask recognition with attention bias}
The original CLIP model can only perform image-level recognition through the \texttt{[CLS]} token. Our work, without changing the parameters of the CLIP model, attempts to allow accurate mask recognition by guiding the attention map of \texttt{[CLS]} token on the region of interests. To achieve this goal, we create a set of \textit{shadow} \texttt{[CLS]} token copies, dubbed \texttt{[SLS]} tokens (MaskCLIP~\cite{ding2022open} adopted similar design, see~\cref{sec:app_sls} for detailed discussion.) These \texttt{[SLS]} tokens are unidirectionally updated by visual tokens, but neither visual tokens nor \texttt{[CLS]} tokens are affected by them (\cref{fig:head}). When updating \texttt{[SLS]} tokens, the predicted attention biases $\mathbf{B}_k \in \mathbb{R}^{h \times w \times N}$ are added to the attention matrix:
\begin{align}
    \mathbf{X}_\texttt{[SLS]}^{l+1} = \texttt{softmax}(\mathbf{Q}_\texttt{[SLS]}^l \mathbf{K}_\texttt{visual}^l + \mathbf{B}_k) \mathbf{V}_\texttt{[SLS]}^l
\end{align}
, where $l$ indicates layer number, $k$ indicates the k-th attention head, $\mathbf{Q}_\texttt{[SLS]} = \mathbf{W_q} \mathbf{X}_{\texttt{[SLS]}}$ and $\mathbf{V}_\texttt{[SLS]} = \mathbf{W_v} \mathbf{X}_{\texttt{[SLS]}}$ are query and value embedding of \texttt{[SLS]} tokens, and $\mathbf{K}_\texttt{visual} = \mathbf{W_k} \mathbf{X}_{\texttt{visual}}$ is the key embedding of visual tokens. $\mathbf{W_q}$, $\mathbf{W_k}$, $\mathbf{W_v}$ are weights of query, key, and value embedding layer, respectively.

We note that the computation complexity here is $\mathcal{O}((T_\texttt{visual} + T_\texttt{[CLS]]} + T_\texttt{[SLS]})^2)$, where $T_\texttt{visual}$, $T_\texttt{[CLS]]}$ and $T_\texttt{[SLS]]}$ are the number of different types of tokens, if implemented by concatenating all types of tokens together and using a masked self-attention layer.  However, we can update the \texttt{[SLS]} token via the cross-attention, which shares the embedding weights with self-attention. Thus the computation complexity becomes $\mathcal{O}((T_\texttt{visual} + T_\texttt{[CLS]]})^2 + T_\texttt{[SLS]} (T_\texttt{visual} + T_\texttt{[CLS]]}))$.

With attention biases, the feature of \texttt{[SLS]} tokens gradually evolves to fit mask prediction, and the class prediction of masks can be easily obtained by comparing the distance/similarity between the \texttt{[SLS]} token and the CLIP text embedding of class names, denoted as $\mathbf{P} \in \mathbb{R}^{C \times N}$, where $C$ is class number. 

\paragraph{Segmentation map generation} With the mask proposals $\mathbf{M} \in \mathbb{R}^{\frac{H}{16} \times \frac{W}{16} \times N}$ and the class prediction of masks $\mathbf{P} \in \mathbb{R}^{C \times N}$, we can compute the segmentation map:
\begin{align}
    \mathbf{S} = \mathbf{M} \times \mathbf{P}^\mathrm{T}
\end{align}
, where $\mathbf{S} \in \mathbb{R}^{\frac{H}{16} \times \frac{W}{16} \times C}$. It is a standard output of semantic segmentation and is therefore compatible with mainstream semantic segmentation evaluation.

To train our model, we follow the practice of \cite{cheng2021mask2former}. The mask generation is supervised with the dice loss $L_{\texttt{mask\_dice}}$ and binary cross-entropy loss $L_{\texttt{mask\_bce}}$. The mask recognition is supervised with the cross-entropy loss $L_{\texttt{cls}}$. The total loss is:
    \begin{equation}
        L_{\texttt{seg}}=\lambda_1 L_{\texttt{mask\_dice}} + \lambda_2 L_{\texttt{mask\_bce}} + \lambda_3 L_{\texttt{cls}}
    \end{equation}
The loss weight $\lambda_1$, $\lambda_2$, $\lambda_3$ are 5.0, 5.0, and 2.0, respectively.
The gradient flow of SAN is shown in \cref{fig:arch}. With end-to-end training, the side adapter network can maximally adapt to the frozen CLIP model, thus the mask proposals and attention biases are \emph{CLIP-aware}.

\section{Experiments}
In this section, we will first introduce the datasets and the evaluation protocol used in our experiments (~\cref{sec:dataset}). Then we will describe the implementation details of our experiments (~\cref{sec:implement}). Finally, we will compare our method with the state-of-art methods (~\cref{sec:system}) and ablate the effectiveness of our method (~\cref{sec:aba}).
\subsection{Dataset and Evaluation Protocol}
\label{sec:dataset}
We conduct experiments on 6 datasets: COCO Stuff~\cite{caesar2018coco}, ADE20K-150~\cite{zhou2017scene}, ADE20K-847~\cite{zhou2017scene}, Pascal Context-59~\cite{mottaghi2014role}, Pascal Context-459~\cite{mottaghi2014role}, and Pascal VOC~\cite{everingham2011pascal}. Following the common practice~\cite{xu2022simple,liang2022open}, all models are trained on the training set of COCO Stuff and evaluated on other datasets.

\paragraph{COCO Stuff} It contains 164K images with 171 annotated classes, which are divided into the training set, the validation set, and the test set containing 118K, 5K, and 41K images, respectively. In our experiments, we use the full 118K training set as the training data by default.

\paragraph{ADE20K-150(ADE-150)} 
It is a large-scale scene understanding dataset with 20K training images and 2K validation images, and a total of 150 annotated classes. 

\paragraph{ADE20K-847(ADE-847)}
It has the same images as ADE20K-150 but more annotated classes (847 classes), which is a challenging dataset for open-vocabulary semantic segmentation.

\paragraph{Pascal VOC(VOC)} 
Pascal VOC contains 20 classes of semantic segmentation annotations, where the training set, and the validation set contain 1464, and 1449 images, respectively.

\paragraph{Pascal Context-59(PC-59)} 
It is a dataset for semantic understanding which contains 5K training images, 5K validation images, and a total of 59 annotated classes.

\paragraph{Pascal Context-459(PC-459)}
It has the same images as Pascal Context-59 but more annotated classes (459 classes), which is also widely used in open-vocabulary semantic segmentation.

\paragraph{Dataset Analysis} 

The relationship between the different datasets is a merely touched problem in the previous paper. To clarify and benefit our understanding of the open-vocabulary ability, we hereby give a straightforward analysis by computing the category similarity between other datasets and the training dataset COCO Stuff. 
We compute the similarity between two datasets with the Hausdorff Distance. For pairwise similarity computing, we extract the text embedding of each concept with the pretrained CLIP text encoder (ViT-L/14) and compute the cosine similarity. The results are presented in~\cref{tab:data}. Among the five validation datasets, Pascal VOC and Pascal Context-59 have a high similarity score of up to 0.9, which means they are better at measuring the \emph{in-domain open-vocabulary ability} in terms of the visual categories. Moreover, Pascal Context-459, ADE20K-150, and ADE20K-847 have a lower similarity score, making them better evaluate the \emph{cross-domain open-vocabulary ability}.
\begin{table}[]
    \centering
    \small
    \begin{tabular}{c|c}
    \toprule
    Dataset     &  Label Sim. to COCO Stuff\\
    \hline
    Pascal VOC     & 0.91 \\
    Pascal Context-59     & 0.86 \\
    Pascal Context-459     & 0.70 \\
    ADE20K-150     & 0.73 \\
    ADE20K-847     & 0.57 \\
    \bottomrule
    \end{tabular}
    \caption{The label-set similarity between validation datasets and training set (\ie COCO Stuff). Measured by Hausdorff distance and cosine similarity based on CLIP text encoder.}
    \label{tab:data}
\end{table}

\paragraph{Evaluation Protocol} Following the common practice~\cite{cheng2021mask2former,xu2022simple,ghiasi2021oseg}, we use the mean of class-wise
intersection over union (mIoU) to measure the performance of our models. For the system-level comparison, we report the mean and variance of 5 trials to ease the randomness. For the ablation study, we only report the average results of 2 trials for saving cost.

\begin{table*}[]
\footnotesize
\centering
\setlength{\tabcolsep}{1.5mm}{
\begin{tabular}{c|c|c|c|c|c|c|c|c}
\toprule
\multirow{1}{*}{Method}                                  & \multirow{1}{*}{VL-Model} & \multirow{1}{*}{Training Dataset} & ensemble. & ADE-847                                & PC-459                                 & ADE-150                                     & PC-59                                     & VOC                                    \\
\hline
Group-VIT~\cite{xu2022groupvit}    & rand. init.                         & CC12M+YFCC                &      no.     & -                                      & -                                       & -                                       & 22.4                                   & 52.3                                   \\ \hline
LSeg+~\cite{ghiasi2021oseg}        & ALIGN RN101              & COCO                                 &      no.    & 2.5                                     & 5.2                                     & 13.0                                    & 36.0                                   & 59.0                                   \\
OpenSeg~\cite{ghiasi2021oseg}      & ALIGN RN101              & COCO                                 &      no.    & 4.0                                     & 6.5                                     & 15.3                                    & 36.9                                   & 60.0                                   \\
LSeg+~\cite{ghiasi2021oseg}        & ALIGN EN-B7              & COCO                                 &      no.    & 3.8                                     & 7.8                                     & 18.0                                    & 46.5                                   & -                                      \\
OpenSeg~\cite{ghiasi2021oseg}      & ALIGN EN-B7              & COCO                                 &      no.    & 6.3                                     & 9.0                                     & 21.1                                    & 42.1                                   & -                                      \\
OpenSeg~\cite{ghiasi2021oseg}      & ALIGN EN-B7              & COCO+Loc. Narr.                      &      no.    & 8.8                                     & 12.2                                    & 28.6                                    & 48.2                                   & 72.2                                   \\ \hline
SimSeg~\cite{xu2022simple}         & CLIP ViT-B/16             & COCO                                &      yes.    & 7.0                                     & 8.7                                     & 20.5                                    & 47.7                                   & 88.4                                   \\
SimSeg\dag                                & CLIP ViT-B/16             & COCO                         &      yes.    & 6.9                                     & 9.7                                     & 21.1                                    & 51.9                                   & 91.8                                   \\
OvSeg~\cite{liang2022open}         & CLIP ViT-B/16             & COCO                                &      yes.    & 7.1                                     & 11.0                                    & 24.8                                    & 53.3                                   & 92.6                                   \\
SAN(ours)                                              & CLIP ViT-B/16             & COCO            &      no.          & $\textbf{10.1}\pm 0.23$   &  $\textbf{12.6}\pm 0.44$         &  $\textbf{27.5}\pm 0.34$          &  $\textbf{53.8}\pm 0.57$          &      $\textbf{94.0}\pm 0.21$                                    \\
\textcolor{gray}{SAN ensemble.}                                             & \textcolor{gray}{CLIP ViT-B/16}             & \textcolor{gray}{COCO}            &      \textcolor{gray}{yes.}            & \textcolor{gray}{${10.7}\pm 0.22$}   &  \textcolor{gray}{${13.7}\pm 0.34$}         &  \textcolor{gray}{${28.9}\pm 0.42$}         &  \textcolor{gray}{${55.4}\pm 0.11$}          &      \textcolor{gray}{${94.6}\pm 0.11$}                                    \\

\hline
MaskCLIP~\cite{ding2022open}       & CLIP ViT-L/14             & COCO                                &       no.   & 8.2                                     & 10.0                                    & 23.7                                    & 45.9                                   & -                                      \\
SimSeg\dag                                & CLIP ViT-L/14             & COCO                         &       yes.   & 7.1                                     & 10.2                                    & 21.7                                    & 52.2                                   & 92.3                                   \\
OvSeg~\cite{liang2022open}         & CLIP ViT-L/14             & COCO                                &       yes.     & 9.0                                     & 12.4                                    & 29.6                                    & 55.7                                   & 94.5                                   \\
SAN(ours)                                              & CLIP ViT-L/14             & COCO            &       no.           & $\textbf{12.4}\pm 0.27$                          & $\textbf{15.7}\pm 0.26$                          & $\textbf{32.1}\pm 0.42$                          & $\textbf{57.7}\pm 0.34$                         & $\textbf{94.6}\pm 0.42$                       \\
\textcolor{gray}{SAN ensemble.}                                              & \textcolor{gray}{CLIP ViT-L/14}             & \textcolor{gray}{COCO}            &       \textcolor{gray}{yes.}          & \textcolor{gray}{${13.7}\pm 0.12$}                          & \textcolor{gray}{${17.1}\pm 0.18$}                          & \textcolor{gray}{${33.3}\pm 0.29$}                          & \textcolor{gray}{${60.2}\pm 0.31$}                         & \textcolor{gray}{${95.5}\pm 0.16$}                         \\ \bottomrule
\end{tabular}
}
\caption{Performance comparison with state-of-the-art methods. \dag~SimSeg~\cite{xu2022simple} trained with a subset of COCO Stuff in their paper. For a fair comparison, we reproduce their method on the full COCO Stuff with their officially released code. * RN101: ResNet-101~\cite{he2016deep}; EN-B7: EfficientNet-B7~\cite{tan2019efficientnet}; SAN ensemble. is the result using ensemble tricks, not the default setting. }
\label{tab:sys_perf}
\end{table*}
\subsection{Implementation Details}
\label{sec:implement}

\paragraph{Training Setting} 
By default, the side adapter network consists of 8 transformer blocks with channel dimensions of 240, attention heads of 6, patch size of 16, and 100 query tokens. 
For ViT-B/16 CLIP model (pretrained on $224^2$ resolution), we used the first 9 blocks for feature fusion and the last 3 blocks for mask recognition, the input resolution is $320^2$. For ViT-L/14 model (pretrained on $336^2$ resolution), we use the first 18 blocks for feature fusion and the last 6 blocks for mask recognition, and the input resolution is $448^2$. For both ViT-B/16 and ViT-L/16, the input resolution of the side-adapter network is $640^2$.

All models are trained on the training set of COCO Stuff dataset. The AdamW optimizer are used with the initial learning rate of 1e-4, weight decay of 1e-4, batch size of 32, and total 60K training iterations, During training, the learning rate is decayed with a poly schedule, with a power of 0.9. We also adopt the data augmentation~\cite{cheng2021mask2former,xu2022simple,liang2022open} with random image resizing in the short-side range of [320,1024] and a crop size of $640^2$. 

\subsection{System level comparison}
\label{sec:system}

\begin{table}[]
\small
\centering

\begin{tabular}{c|c|c|c}
\toprule
Method  & Param. (M) & GFLOPs & FPS                   \\
\hline
SimSeg & 61.1     & 1916.7 & 0.8                   \\
OvSeg* &147.2  &1916.7 & 0.8                   \\
MaskCLIP*& 63.1&307.8& 4.1 \\
SAN(ours)& \textbf{8.4}       &\textbf{64.3}  & \textbf{15.2}    \\
\bottomrule
\end{tabular}
\caption{Training and testing efficiency comparison with other methods. \emph{Param.} stands for the total number of trainable parameters in the methods in millions. The input image is of $640\times 640$ resolution. And the clip model is ViT-B/16. * no official code available yet and we re-implement their methods following the description in their papers.  OvSeg~\cite{liang2022open} has similar structures to SimSeg~\cite{xu2022simple} but it finetuned the whole CLIP model,resulting in much more trainable parameters.}
\label{tab:sys_resource}
\end{table}

In~\cref{tab:sys_perf}, we compare our method with other state-of-the-art methods. In comparison with other methods that also use the CLIP ViT models and COCO Stuff dataset, without using ensemble trick\footnote{Previous works~\cite{xu2022simple,liang2022open} ensemble the predictions of the model fine-tuned on COCO Stuff with the predictions of frozen CLIP to get better performance. In our approach, we ensemble our model with the model fine-tuned on COCO Stuff.}, our method surpasses other methods under the same setting with an average of +1.8 mIoU for CLIP ViT-B/16, and an average of +2.3 mIou for ViT-L/14, respectively. Further applying the ensemble trick increases the gap to an average of +2.9 mIoU and +3.7 mIoU for CLIP VIT-B/16 and ViT-L/14, respectively.

Notably, the improvements of our method are more pronounced on the ADE-847. As we discussed in \cref{tab:data}, ADE-847 has fewer similar classes to COCO-Stuff, and we argue that the better performance on ADE-847 further affirms the stronger open-vocabulary recognition capability of our approach.

Furthermore, we compare with other methods: SimSeg~\cite{xu2022simple}, OvSeg~\cite{liang2022open}, and MaskCLIP~\cite{ding2022open}, which also use CLIP ViT models, in terms of trainable parameters, GFLOPs and inference time (FPS). For a fair comparison, we test all methods under the same environment: single Titan Xp GPU, Xeon E5 v2 CPU (32 core), 252G RAM, PyTorch 1.9.0, and CUDA 11.3. We use images of $640^2$ resolution for all models, and process a single image per inference. The results are summarized in~\cref{tab:sys_resource}. Our approach outperforms other methods in all aspects. 

\begin{figure*}
    \centering
    \includegraphics[width=\textwidth]{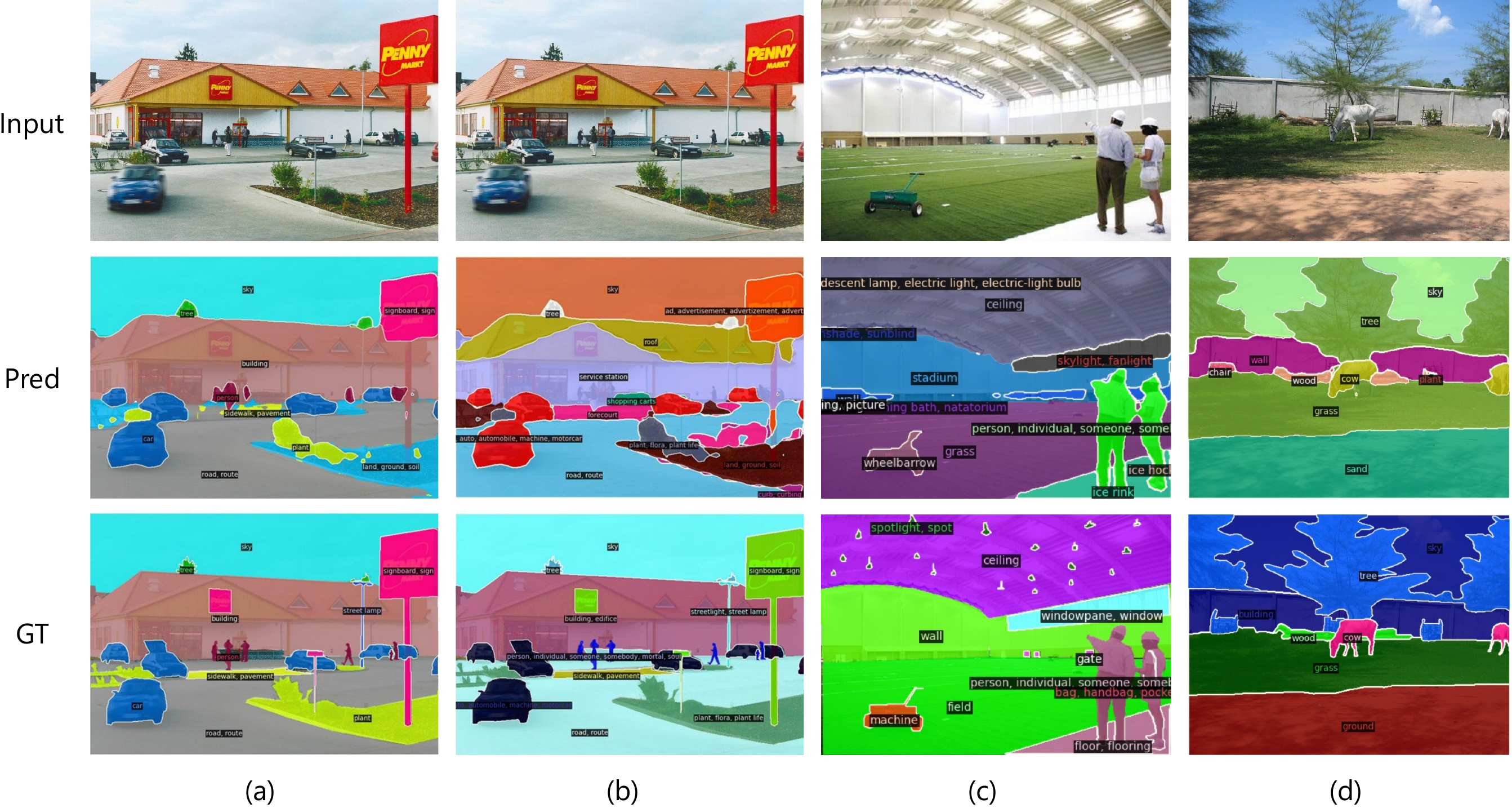}
    \caption{Qualitative results of our method. (a) and (b) are the results of the same input images with different vocabularies (ADE-150 and ADE-847 respectively).}
    \label{fig:seg_vis}
\end{figure*}

We also visualize the the predictions with our best ViT-L/14 model in~\cref{fig:seg_vis}. ~\cref{fig:seg_vis} (a) and (b) are of the same input image but with different vocabularies (from ADE-150, ADE-847 respectively), and e.g., the \emph{signboard, sign} is correctly classified in ADE-150 but is mis-classified as~\emph{ad, advertisement} in ADE-847. And we assume that the model is not good at verifying the difference between \emph{advertisement} and \emph{signboard}.

\subsection{Ablation Studies}
\label{sec:aba}
We ablate the key design choices of our method on ADE-150. If not specified, the ViT-B/16 CLIP model and an 8-layer side adapter network with a feature dimension of 240 and an attention head of 6 are used as the default setting.

\begin{table}[]
    \centering
    \small
    \begin{tabular}{c|c|c}
    \toprule
        Description. & Layers & mIoU \\
        \hline
        w/o. fusion & none & 21.1 \\
        \hline
        \multirow{3}{*}{single-fusion} & stem & 20.0\\ 
         &  3rd layer & 24.1\\
        &  6th layer  & 26.2\\
        &  9th layer &  27.1 \\
         \hline
        \multirow{3}{*}{multi-fusion} & \{6,9\}-layers & 27.0  \\
        & \{3,6,9\}-layers & 27.7   \\
        &  \{stem,3,6,9\}-layers & \textbf{27.8} \\
    \bottomrule
    \end{tabular}
    \caption{Different feature fusion strategies. The last 3 layers of ViT-B/16 are used for mask prediction in all experiments.}
    \label{tab:aba_fusion}
\end{table}

\begin{table}[]
    \centering
    \small
    \begin{tabular}{c|c|c}
    \toprule
        \#Feature Fusion Layers & \#Recognition Layers & mIoU \\
        \hline
        12 & 12 & 27.6 \\
        \hline
         11 & 1 & 25.9\\ 
         10 & 2&27.3 \\
         9 & 3 & \textbf{27.8}   \\
         6&6&   26.9   \\
         3&9 &  23.8     \\
    \bottomrule
    \end{tabular}
    \caption{The trade-off between the number of feature fusion layers and the number of mask prediction layers. \emph{Note}: the 2nd row (\ie the \{12,12\} setting) is the \emph{twice-forward} baseline.}
    \label{tab:clip_split}
\end{table}
 
\paragraph{Importance of feature fusion.}
The key to SAN being lightweight is leveraging the strong features of the CLIP model. We experimentally illustrate the importance of feature fusion in~\cref{tab:aba_fusion}. 
Without fusing the CLIP feature, the mIoU would drop from 27.8 to 21.1. In addition, we also noticed that fusing the feature of deeper layers (\eg 9th layer) is better than fusing shallower layers (\eg stem layer), and only fusing the feature of 9th layer can reach 27.1 mIoU, which is +6.0 mIoU higher than baseline without feature fusion. This observation is consistent with the intuition that deeper features tend to be more semantic. In addition, fusing features from multiple layers can further improve performance compared to single-layer fusion by +0.8 mIoU.

To minimize the inference cost of CLIP, we adopt a \emph{single-forward} design that the shallower layers are used for feature fusion, and other deeper layers are used for mask recognition, and thus a trade-off is required, which is examined in \cref{tab:clip_split}, and the best performance is achieved when the first 9 layers are used for feature fusion and the last 3 layers for mask recognition. In addition, we also compared with the \emph{twice-forward} baseline (2nd row in \cref{tab:clip_split}) and did not find a significant difference. \footnote{We note a 0.2 mIoU gap which could arise from randomness.}

\paragraph{Importance of \emph{CLIP-aware} mask prediction.}
\begin{figure}
    \centering
    \includegraphics[width=0.47\textwidth]{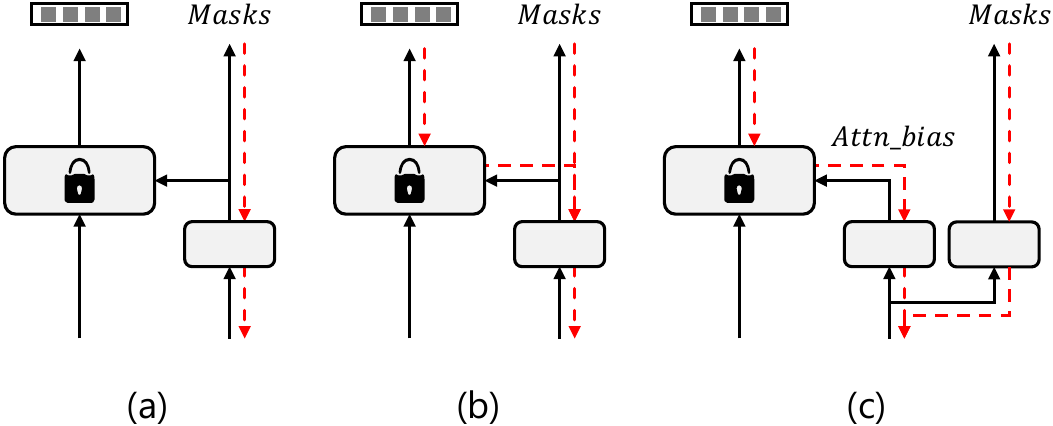}
    \caption{Design choice of mask prediction head. (a) Two-stage training with single head and blocking gradients from CLIP. (b) End-to-end training with single head (c) End-to-end training with decoupled head. The red dotted line indicates the gradient flow during training.}
    \label{fig:dis_head}
\end{figure}
\begin{table}[]
\small
\centering
\begin{tabular}{c|c|c|c}
\toprule
Description & Backbone & \emph{CLIP-aware} & mIoU \\
\hline
\textcolor{gray}{SimSeg} & \textcolor{gray}{ViT-B/16} & \textcolor{gray}{no.} & \textcolor{gray}{21.1} \\
\textcolor{gray}{MaskCLIP} & \textcolor{gray}{ViT-L/14} & \textcolor{gray}{no.} & \textcolor{gray}{23.7} \\
\hline
two-stage training  & ViT-B/16 &  no. &   21.6 \\
e2e training & ViT-B/16 & yes.  & \textbf{26.1} (+4.5) \\
\bottomrule
\end{tabular}
\caption{\emph{Two-stage} \vs \emph{end-to-end}. The significant improvement proves the importance of \emph{CLIP-aware} mask prediction. }
\label{tab:aba_clip_aware}
\end{table}

Unlike other two-stage frameworks~\cite{xu2022simple,ding2022open}, our approach is an end-to-end training framework. We study the difference between the two frameworks in~\cref{tab:aba_clip_aware}.
As the attention bias branch must be trained through CLIP, for comparison, we use the mask proposals instead of the attention bias in the self-attention layers of CLIP. If the gradients from CLIP are blocked, the method degenerates into a two-stage framework, \ie, the mask prediction is isolated from CLIP recognition. Otherwise, the method is a \emph{single} head end-to-end training framework, and the mask prediction is \emph{CLIP-aware}.

\cref{tab:aba_clip_aware} shows the end-to-end training has \emph{+4.5} mIoU improvements over two-stage baseline. Besides, we list the results of other two-stage methods as a reference, showing that our two-stage baseline can achieve reasonable performance.

\paragraph{Importance of the \emph{decoupled} head.} 

\begin{table}[]
\small
\centering
\begin{tabular}{c|c|c}
\toprule
Head & E2E Training & mIoU  \\
\hline
\emph{single} head  & yes. & 26.1  \\
\emph{decoupled} head & yes. &  \textbf{27.8} (+1.7)   \\
\bottomrule
\end{tabular}
\caption{Comparison on \emph{single} head and \emph{decoupled} head. With few additional parameters and flops, \emph{decoupled} head improves a notable performance. All models are trained end-to-end.}
\label{tab:aba_decouple}
\end{table}
 We study the effect of the \emph{decoupled} head design in~\cref{tab:aba_decouple}. Compared with the \emph{single} head model, the \emph{decoupled} head model has \emph{+1.7} mIoU improvements. Note that both models are trained end-to-end, so their mask predictions are all \emph{CLIP-aware}.

\paragraph{Asymmetric input resolution.} 
\begin{table}[]
\small
\centering

\begin{tabular}{c|c|c}
\toprule
Resolution. & GFLOPs & mIoU \\
\hline
$192^2$     &  39.4  &  25.3        \\
$224^2$      &    44.3       &   26.3   \\
$320^2$      &   64.3        &  \textbf{27.8}    \\
$448^2$ &   106.3  &   26.1     \\
$640^2$      &  213.4  &   24.6         \\
\bottomrule
\end{tabular}
\caption{The influence of ViT-B/16 CLIP model input resolution. We vary CLIP input resolutions, while always using $640^2$ images in the side-adapter network.}
\label{tab:input_res}
\end{table}
\begin{table}[]
\small
\centering

\begin{tabular}{c|c|c}
\toprule
Description. & Resolution. & mIoU \\
\hline
fixed pos embed. &$320^2$ &27.0 \\
ft. pos embed. &$320^2$ & \textbf{27.8} \\
\bottomrule
\end{tabular}
\caption{Fine-tuning the position embedding can improve the performance.}
\label{tab:aba_finetune_ape}
\vspace{-1em}
\end{table}
We based on the officially released ViT CLIP models. They are designed for low-resolution input images (\eg $224^2$), while semantic segmentation requires high-resolution images. To resolve the conflicts on input resolution, we use low-resolution images for CLIP model and high-resolution images for SAN model. \cref{tab:input_res} shows how the different image resolutions of CLIP model affect performance. 
In addition, by default, we fine-tune the position embedding of CLIP model, its effects are shown in~\cref{tab:aba_finetune_ape}.

\begin{table}[]
\small
\centering

\begin{tabular}{c|c|c|c}
\toprule
Width of SAN & Param. (M) & GFLOPs&mIoU  \\
\hline
144       &    4.2           & 53.6     &    26.7 \\
192       &    6.1           & 58.6     &   27.4    \\
240       &    8.4           & 64.3     &   \textbf{27.8}    \\
288       &    11.1           &   70.9   &   27.3   \\
\bottomrule
\end{tabular}
\caption{The influence of capacity of SAN. \emph{Param.} stands for the total number of trainable parameters in the model in millions.}
\label{tab:dis_cap}
\vspace{-1em}
\end{table}
\paragraph{Discussion on the Parameter Efficiency}
\cref{tab:dis_cap} examines how the capacity of the side adapter network affects the performance. We find that small models can already achieve good performance, while larger model capacity does not provide significant gains. We speculate that this is because our approach has taken advantage of the features of CLIP, thus relieving the need for a large model capacity.

\section{Conclusion}
In this work, we presented the SAN framework for open-vocabulary semantic segmentation. Our framework succeeds in leveraging the features of the frozen CLIP model and an end-to-end pipeline to adopt the frozen CLIP model maximally. Notably, the proposed framework significantly outperforms the previous state-of-the-art methods on five semantic segmentation benchmarks with much fewer trainable parameters and much less computation cost.

\appendix

\section{More Design Choices}
\subsection{Attention Bias Format}
In~\cref{tab:attn_aba_per_head_layer}, we further ablate the format of the attention bias. Using different biases for each attention head achieves better performance, but further introducing layer-wise attention bias bring no gains.

\begin{table}[]
\small
\centering
\vspace{-1em}
\begin{tabular}{c|c|c}
\toprule
Per Head?&Per Layer?& mIoU \\
\hline
no.&no. &26.2\\
yes.&no. &\textbf{27.8} \\
no.&yes. & 26.3\\
yes.&yes. & 27.4\\
\bottomrule
\end{tabular}
\caption{Ablation on the design of attention bias.}
\label{tab:attn_aba_per_head_layer}
\end{table}

\begin{table}[]
    \small
       \centering
    
    \begin{tabular}{c|c}
        \toprule
        Init. method & mIoU \\
        \hline
        Learned embedding &  27.2 \\
        \texttt{[CLS]} & \textbf{27.8} \\
         \bottomrule
    \end{tabular}
    \caption{Initialization method of \texttt{[SLS]} token.}
    \label{tab:sls}
\end{table}

\subsection{ [SLS] tokens.}
\label{sec:app_sls}
As discussed in the paper, we use the copies of \texttt{[CLS]} token as our \texttt{[SLS]} tokens. The \texttt{[SLS]} tokens are conceptually similar to the Mask Class Tokens (MCT)~\cite{ding2022open} but differ in their implementation details, such as how they are updated and where they are introduced in the CLIP model. Furthermore, while \texttt{[SLS]} tokens are initialized from the \texttt{[CLS]} token by default, ~\cref{tab:sls} shows that using the learned embedding is only marginally worse.

\subsection{Prompt Engineering}

Prompt engineering has been proven useful for open-vocabulary semantic segmentation. Following the common practice~\cite{xu2022simple,liang2022open}, we use multiple templates to decorate the class names and average their text embeddings as the final used text embedding for each class in inference. The templates are listed in~\cref{tab:prompt}, and the effects of prompt engineering are shown in~\cref{tab:prompt_effect}, which can improve 1.2 mIoU, 0.7 mIoU on ADE-150 and ADE-847, respectively.

\begin{table}[h]
    \small
    \centering
    \begin{tabular}{l}
    \toprule
    ``a photo of a \{\}.", \\
    ``This is a photo of a \{\}",\\
    ``There is a \{\} in the scene",\\
    ``There is the \{\} in the scene",\\
    ``a photo of a \{\} in the scene",\\
    ``a photo of a small \{\}.",\\
    ``a photo of a medium \{\}.",\\
    ``a photo of a large \{\}.",\\
    ``This is a photo of a small \{\}.",\\
    ``This is a photo of a medium \{\}.",\\
    ``This is a photo of a large \{\}.",\\
    ``There is a small \{\} in the scene.",\\
    ``There is a medium \{\} in the scene.",\\
    ``There is a large \{\} in the scene.",\\
    \bottomrule
    \end{tabular}
    \caption{Prompt templates used in our method.}
    \label{tab:prompt}
\end{table}

\begin{table}[]
    \small
    \centering
    \begin{tabular}{c|c|c|c}
    \toprule
        method & w/ prompt. & ADE-150&ADE-847 \\
        \hline
        SimSeg~\cite{xu2022simple} & yes. & 21.1&6.9 \\
        OvSeg~\cite{liang2022open} & yes. & 24.8&7.1 \\ 
        \hline
        SAN & no. & 26.6&9.5 \\ 
        SAN   & yes. & \textbf{27.8} & \textbf{10.2} \\ 
    \bottomrule
    \end{tabular}
    \caption{Effects of prompt engineering. The single template \emph{``a photo of \{\}."} is used for models without using prompt engineering.}
    \label{tab:prompt_effect}
\end{table}

\section{Comparison with fine-tuned models} 
\begin{figure}
    \centering
    \includegraphics[width=0.43\textwidth]{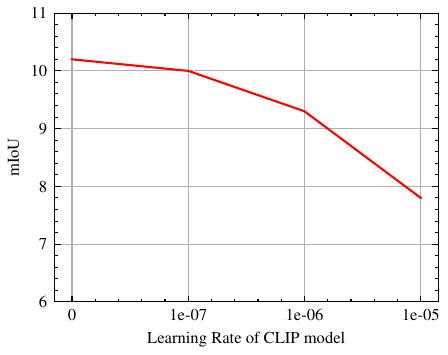}
    \caption{Performance on ADE-847 when fine-tuning the CLIP model with different learning rates. The learning rate of 0 is the \emph{frozen} CLIP model.}
    \label{fig:lr2res}
\end{figure}

We study the effects of fine-tuning the pre-trained CLIP model in ADE-847 dataset based on the side adapter network. For a fair comparison, we applied different initial learning rates to the CLIP model while using the same initial learning rate of the side adapter network (\ie 1e-4) and keeping other hyper-parameters unchanged. The results are shown in~\cref{fig:lr2res}. The learning rate of 0 indicates \emph{frozen} CLIP model. As the learning rate applied in CLIP mode increases, models perform worse in ADE-847 (\textit{but perform better on COCO Stuff dataset}), indicating that the open-vocabulary capability of the CLIP model is disrupted.

{\small
\bibliographystyle{ieee_fullname}
\bibliography{egbib}
}

\end{document}